\documentclass{llncs}

\usepackage{graphicx}
\usepackage[compatibility=false]{caption}
\usepackage{subcaption}
\usepackage[usenames, dvipsnames]{color}

\begin{document}
\frontmatter          
\pagestyle{empty}  
\mainmatter              
\title{CT Image Denoising with Perceptive Deep Neural Networks}
\titlerunning{Short Title}  
\author{Qingsong Yang\inst{1} \and Pingkun Yan\inst{2} \and Mannudeep K. Kalra\inst{3} \and Ge Wang\inst{1}}
\authorrunning{Anonymous et al.}   
%
\tocauthor{author2 (Affiliation of Author2), author3 (Affiliation of
Author3), }
\institute{Department of Biomedical Engineering, Rensselaer Polytechnic Institute, Troy, NY, USA
  \and
  Philips Research North America, Bethesda, MD, USA
  \and
Divisions of Thoracic and Cardiac Imaging, Department of Imaging, Massachusetts General Hospital, Harvard Medical School, Boston, MA, USA
}


\maketitle              

\begin{abstract}

Increasing use of CT in modern medical practice has raised concerns over associated radiation dose. Reduction of radiation dose associated with CT can increase noise and artifacts, which can adversely affect diagnostic confidence. 
Denoising of low-dose CT images on the other hand can help improve diagnostic confidence, which however is a challenging problem due to its ill-posed nature, since one noisy image patch may correspond to many different output patches. In the past decade, machine learning based approaches have made quite impressive progress in this direction. However, most of those methods, including the recently popularized deep learning techniques, aim for minimizing mean-squared-error (MSE) between a denoised CT image and the ground truth, which results in losing important structural details due to over-smoothing, although the PSNR based performance measure looks great. In this work, we introduce a new perceptual similarity measure as the objective function for a deep convolutional neural network to facilitate CT image denoising. Instead of directly computing MSE for pixel-to-pixel intensity loss, we compare the perceptual features of a denoised output against those of the ground truth in a feature space. Therefore, our proposed method is capable of not only reducing the image noise levels, but also keeping the critical structural information at the same time. Promising results have been obtained in our experiments with a large number of CT images.
\end{abstract}

\section{Introduction}
\label{sec:intro}

X-ray computed tomography (CT) is a critical medical imaging tool in modern hospitals and clinics. However, the potential radiation risk has attracted increasingly more public concerns on the use of x-ray CT \cite{brenner2007computed,de2004risk}. Lowering the radiation dose tends to significantly increase the noise and artifacts in the reconstructed images, which can compromise diagnostic information. To reduce noise and suppress artifacts in low-dose CT images, extensive efforts were made via image post-processing. For example, the non-local means (NLM) method was adapted for CT image denoising \cite{ma2011low}. 
Based on the compressed sensing theory, an adapted K-SVD method was proposed in \cite{chen2013improving} to reduce artifacts in CT images. Moreover, the block-matching 3D (BM3D) algorithm was used for image restoration in several CT imaging tasks \cite{feruglio2010block}. 
Image quality improvement was clearly demonstrated in those applications, however, over-smoothness and/or residual errors were also observed in the processed images. Despite these efforts, CT image denoising remains challenging because of the non-uniform distribution of CT imaging noise.

With the recent explosive development of deep neural networks, researchers tried to tackle this denoising problem through deep learning. Dong et al. \cite{DongLHT16} developed a convolutional neural network (CNN) for image super-resolution and demonstrated a significant performance improvement compared with other traditional methods. The work was then adapted for low-dose CT image denoising \cite{Chen1610.00321}, where similar performance gain was obtained.
However, over-smoothing remains a problem in the denoised images, where important textural clues were often lost. The root cause of the problem is the image reconstruction error measurement used in all the learning based methods. As revealed by the recent research \cite{Johnson1603,Ledig1609}, using the per-pixel mean squared error (MSE) between the recovered image and the ground truth as the reconstruction loss to define objective function results in over-smoothness and lacking of details. As an algorithm tries to minimize per-pixel MSE, it overlooks any image features critical for human perception.

In this paper, we propose a new method for CT image denoising by designing a perceptive deep CNN that relies on a perceptual loss as the objective function. During our research, it was drawn to our attention that minimizing MSE between the denoised CT image and the ground truth leads to the loss of important details, although the peak signal to noise ratio (PSNR) based evaluation numbers are excellent. That is because PSNR is equivalent to the per-pixel Euclidean intensity difference. Therefore, a model maximizing PSNR after successful training always achieves very high PSNR values. However, the perceptual evaluation of the denoised images generated by such a model is not necessarily better than that of the original noisy images from experts' point of view.

In our proposed method, instead of directly computing MSE summarizing pixel-to-pixel intensity differences, we compare the denoised output against the ground truth in another high-dimensional feature space, achieving denoising and keeping critical structures at the same time. We introduce a new perceptual similarity as the objective function of the CNN for CT image denoising. The rationale behind our work is two-fold.
First, when human compares two images, the perception is not performed pixel-by-pixel. Human vision actually extracts features from images and compare them \cite{Nixon2008}. Therefore, instead of using pixel-wise MSE, we employ another pre-trained deep CNN (the famous VGG) for feature extraction and compare the denoised output against the ground truth in terms of the extracted features. 
Second, from a mathematical point of view, CT images are not uniformly distributed in a high-dimensional Euclidean space. They reside more likely in some low-dimensional manifold. With MSE, we are not measuring the intrinsic similarity between the images, but just their superficial differences, i.e., the Euclidean distance. However, by comparing images using extracted features, we actually project the them onto a manifold and calculate the geodesic distance therein. By measuring the intrinsic similarity between images, our proposed approach can produce results with not only lower noise but also sharper details.

The rest of this paper is organized as follows. The details of our proposed method are given in Section~\ref{sec:method}. The performed experiments and discussions are presented in Section~\ref{sec:exp}. Finally, Section~\ref{sec:conclusions} draws conclusions and discusses our future work.

\section{Method}
\label{sec:method}

In this section, we first present the loss functions that we use for measuring the image reconstruction error. The proposed denoising deep network is then described.

\subsection{Loss Functions}

Our proposed method defines the objective loss function of the denoising CNN using feature descriptors. Let $\{ \phi_i(I) | i=1,\ldots,N \}$ denote $N$ different feature maps of an image $I$. Each map has the size of $h\times w \times d$, where $h$, $w$ and $d$ denote height, width and depth, respectively. The feature reconstruction loss can then be defined as
\begin{equation}
\mathcal{L}_{\phi_i}(\hat{I}, I_{gt}) 
= \frac{1}{hwd}\|\phi_i(\hat{I}) - \phi_i(I_{gt}) \|^2,
\label{eq:loss}
\end{equation}
where $\hat{I}$ and $I_{gt}$ are the denoised image and corresponding ground truth, respectively.
In our work, the well-known pre-trained VGG network \cite{VGG} has been used for feature extraction. Although VGG was originally trained for natural image classification, technical analysis shows that many feature descriptors learned by VGG are quite meaningful for human \cite{Mahendran2016}, which suggests that it also learns general perceptual features not specific to any particular kind of images.

\subsection{Network Architecture}

\begin{figure}[t]
	\centering
	\includegraphics[width=\textwidth]{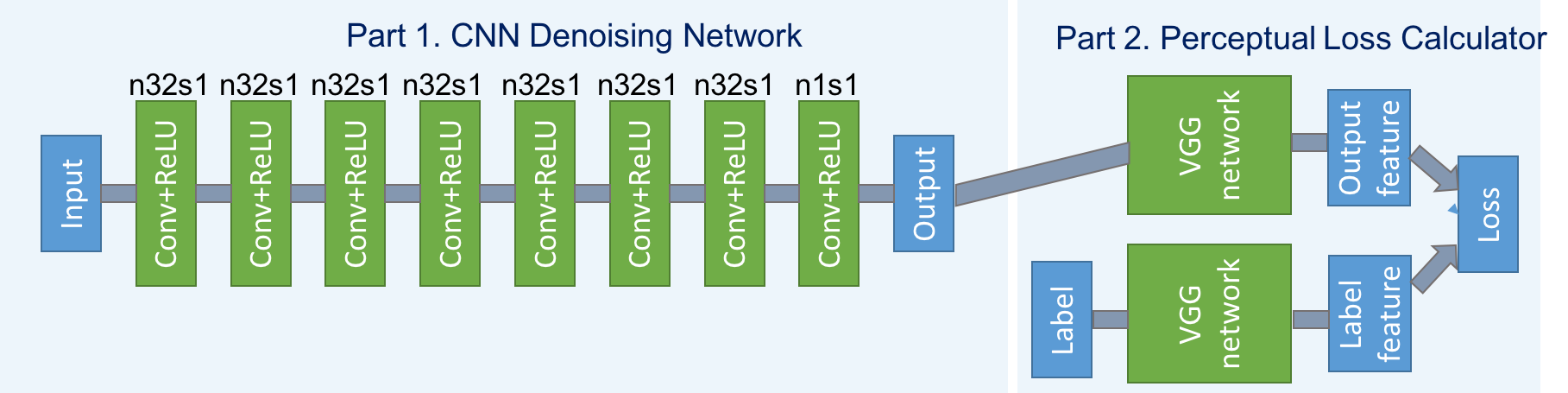}
	\caption{Proposed network structure.}
	\label{fig:net_str}
\end{figure}

Our developed network consists of two parts, the CNN denoising network and the perceptual loss calculator, as shown in Fig.~\ref{fig:net_str}. To learn denoising images containing different structures and intensities, a deep enough network is required to handle the sophistication. In our work, the CNN denoising network was constructed by 8 convolutional layers. Following the common practice in the deep learning community \cite{taxonomy2016}, small $3\times3$ kernels were used in each convolutional layer. Due to the stacking structure, such a network can cover a large enough receptive field efficiently. Each of the first 7 hidden layers of the denoising network had 32 filters. The last layer generates only one feature map with a single $3\times3$ filter, which is also the output of our denoising network. 
We used Rectified Linear Unit (ReLU) as the non-linear activation function for the 7 hidden layers.

The second part of the network is the perceptual loss calculator, which is realized by using the pre-trained VGG network \cite{VGG}. A denoised output image $\hat{I}$ from the first part and the ground truth image $I_{gt}$ are fed into the pre-trained VGG network for feature extraction. Then, the objective loss is computed using the extracted features from a specified layer according to Eqn.~(\ref{eq:loss}). The reconstruction error is then back-propagated to update the weights of the CNN network only, while keeping the VGG parameters intact. 

The VGG network has 16 convolutional layers, each followed by a ReLU layer and 4 pooling layers. In our experiment, we tested the feature maps generated at the first ReLU layer before the first pooling layer, named relu1\_1, and the first and fourth ReLU layers before the third pooling layer, named relu3\_1 and relu3\_4, respectively. The corresponding networks are referred to as CNN-VGG11, CNN-VGG31, and CNN-VGG34 respectively.

\section{Experiments}
\label{sec:exp}


\subsection{Materials and Network Training}

In our work, we trained all the networks on a NVIDIA GTX980 GPU using random samples from the cadaver CT image dataset collected at Massachusetts General Hospital (MGH) \cite{MGH_DATA}. These cadavers were repeatedly scanned under a GE Discovery 750 HD scanner at different noise levels, with the noise index (NI) values of 10, 20, 30, and 40 respectively. In addition, the projection data were used for CT image reconstruction with two different methods. While one is the classic filtered back-projection (FBP) method, the other is a model-based fully iterative reconstruction (MBIR) vendor-specific technique named VEO (GE Healthcare, Waukesha, WI). The MBIR technique has a strong capability of noise suppressing, but the traditional FBP method does not. In our experiment, we used FBP reconstruction from 30NI dataset (high noise level) as the network input and the corresponding VEO reconstruction from 10NI dataset (low noise level) as the ground truth images. 

The proposed network was implemented and trained using the Caffe toolbox \cite{caffe}. At the training phrase, we randomly extracted and selected 100,000 image patches of size $32\times32$ from 2,600 CT images. We first trained a CNN with the same structure as shown in Fig.~\ref{fig:net_str} but using the mean-square-error (MSE) loss, which is named CNN-MSE. The network was trained for 1,920 epochs. Then, the CNN-MSE weights were used to initialize the CNN-VGG11, CNN-VGG31, and CNN-VGG34 networks. In our experiments, we noticed that the new networks can be trained very quickly. In some cases, only 10 epochs were enough to obtain good results, and further training did not help much.

\subsection{Experimental Results}

At the validation stage, whole CT images were used as input. We tested the networks using 500 images from two cadavers' whole body scan. For comparison, we also tested the classic BM3D method \cite{dabov2009bm3d} and the recent work on SRCNN \cite{Chen1610.00321,DongLHT16} named as CNN-MSE. 

\begin{figure}[t]
	\centering
	\includegraphics[width=\textwidth]{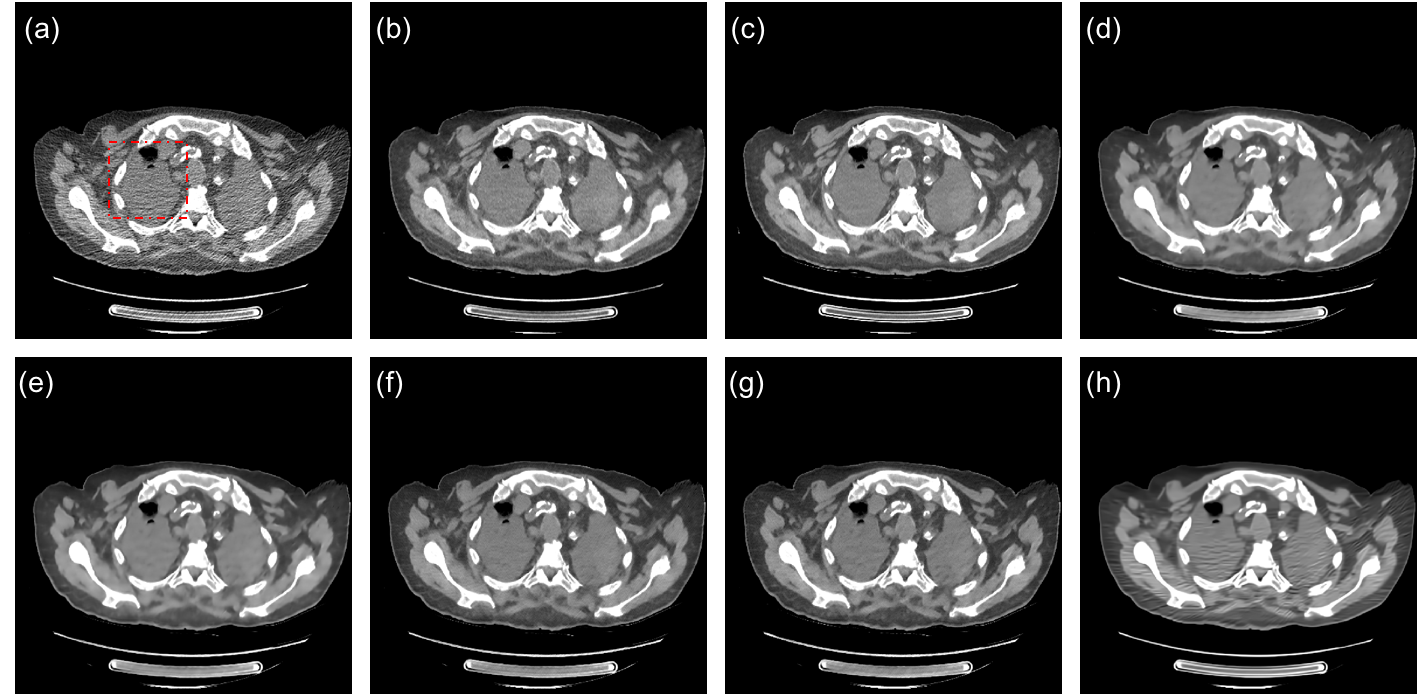}
	\caption{Image denoising example with (a) the input FBP30NI image, (b) VEO30NI, (c) the ground truth of VEO10NI, and restored results from (d) CNN-MSE, (e) CNN-VGG11, (f) CNN-VGG31, (g) CNN-VGG34, (h) BM3D. The display window is [-160, 240]HU.}
	\label{fig:example1}
\end{figure}

\begin{figure}[t]
	\centering
	\includegraphics[width=\textwidth]{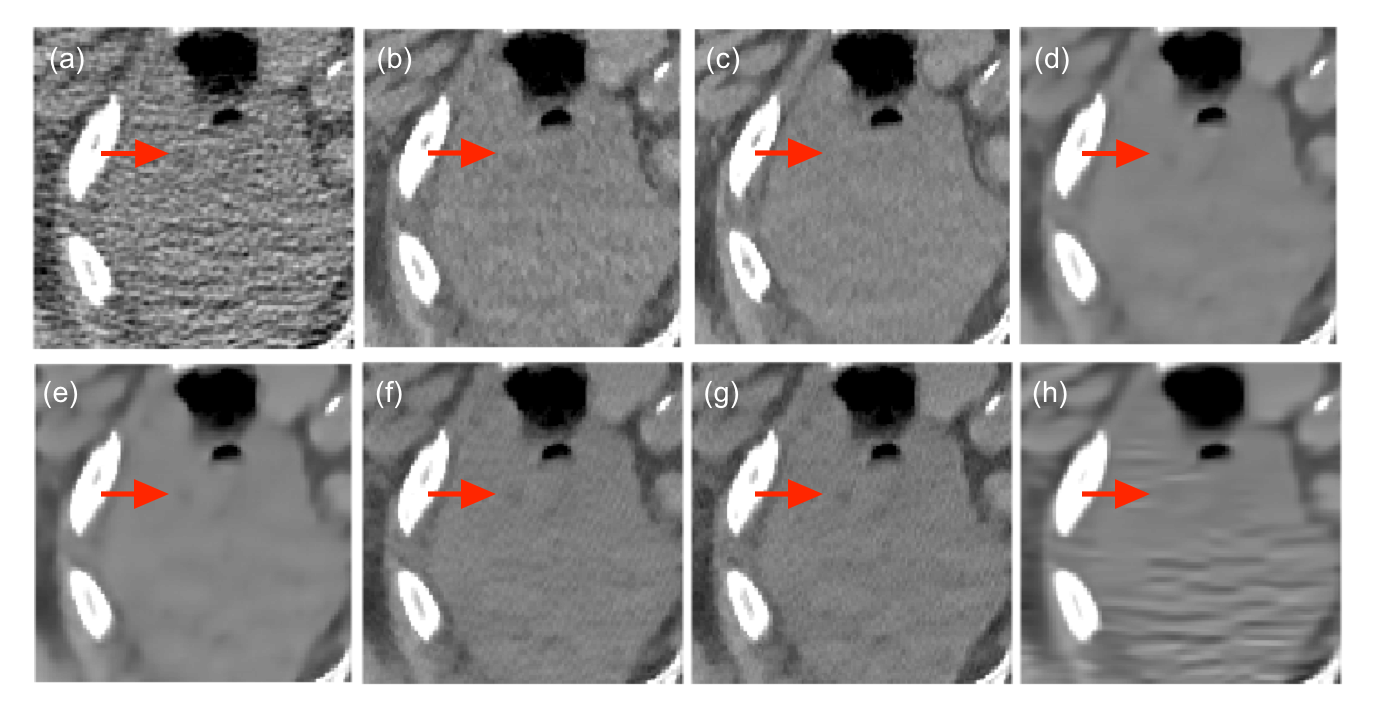}
	\caption{Zoomed ROI marked in Fig.~\ref{fig:example1}. (a) FBP30NI, (b) VEO30NI, (c) VEO10NI, (d) CNN-MSE, (e) CNN-VGG11, (f) CNN-VGG31, (g) CNN-VGG34, (h) BM3D}
	\label{fig:ROI1}
\end{figure}

\begin{figure}[t]
	\centering
	\includegraphics[width=\textwidth]{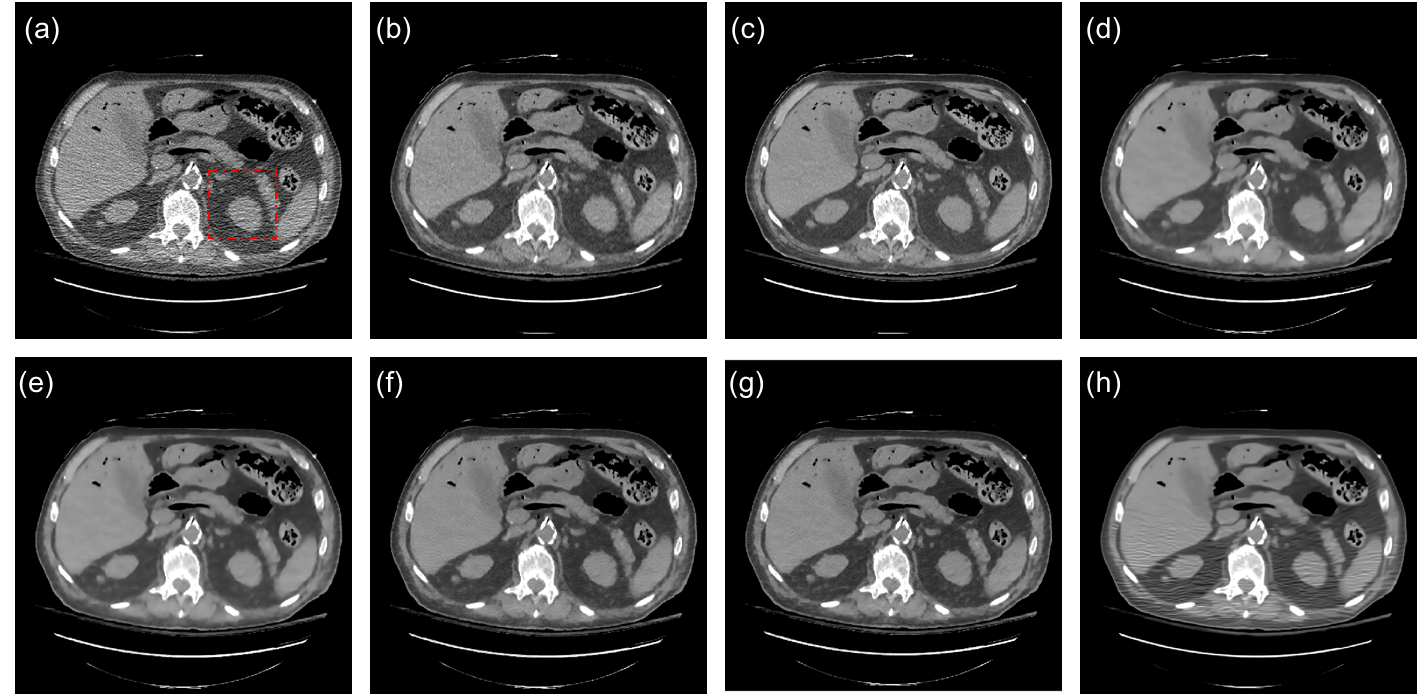}
	\caption{Second set of recovered images in comparison with the original images (a) FBP30NI and (b) VEO30NI, ground truth (c) VEO10NI, and restored images from (d) CNN-MSE, (e) CNN-VGG11, (f) CNN-VGG31, (g) CNN-VGG34, and (h) BM3D. The display window is [-160, 240]HU.}
	\label{fig:example2}
\end{figure}

\begin{figure}[t]
	\centering
	\includegraphics[width=\textwidth]{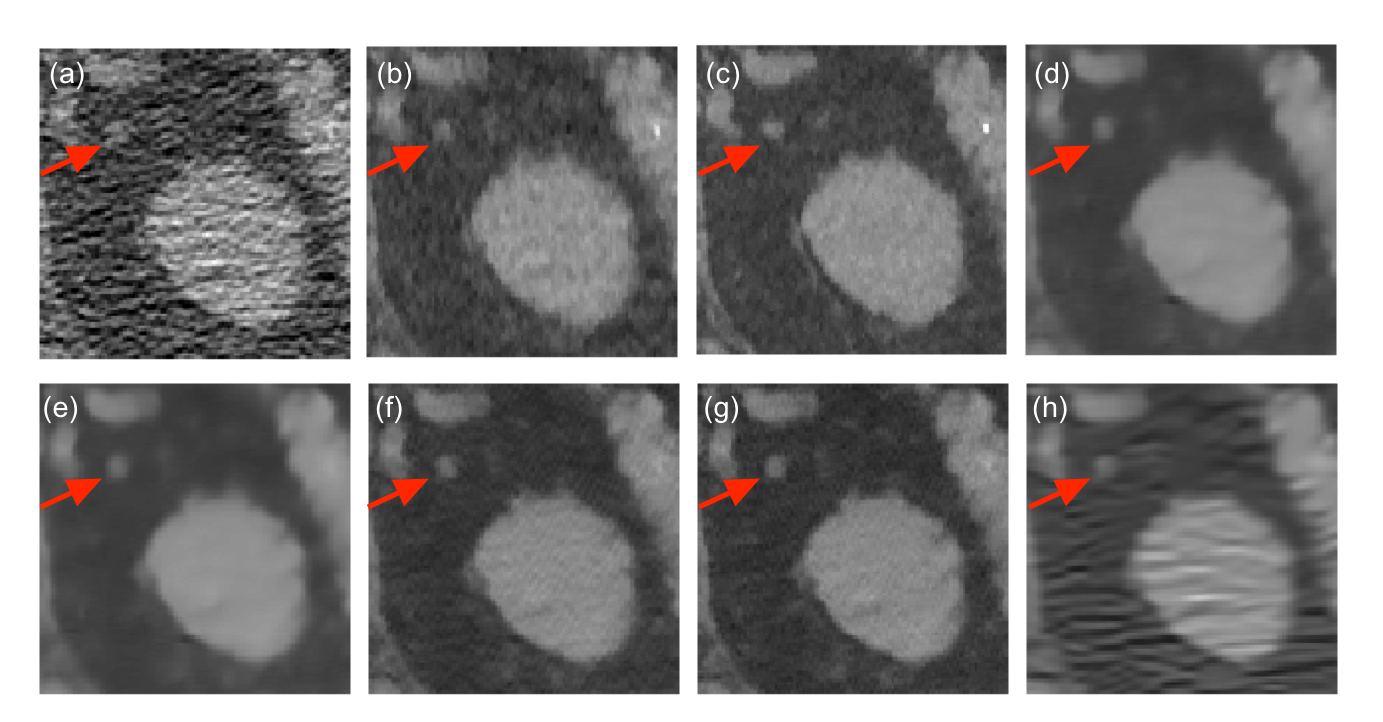}
	\caption{Zoomed ROI in Fig. \ref{fig:example2}. (a) FBP30NI, (b) VEO30NI, (c) VEO10NI, (d) CNN-MSE, (e) CNN-VGG11, (f) CNN-VGG31, (g) CNN-VGG34, (h) BM3D}
	\label{fig:ROI2}
\end{figure}

Figs.~\ref{fig:example1} and \ref{fig:example2} show two examples of the denoised images. To make the differences clearer, ROIs indicated in the red rectangular areas in those figures are zoomed and shown in Figs.~\ref{fig:ROI1} and \ref{fig:ROI2}, respectively. From these images, it is seen that the images recovered by CNN-MSE and CNN-VGG11 got over-smoothed with some details missing. On the contrary, CNN-VGG31 and CNN-VGG34 yielded images of better contrast and more similar to the VEO images. As for BM3D, it gave different visual effects on different images. In Fig. \ref{fig:ROI1}(h), the nodule pointed by the red arrow was smoothed out, while the streak artifacts were reserved in Fig. \ref{fig:ROI2}(h). This can be explained by the non-uniformity of image noise.
%
In addition, although the low contrast lesions (pointed by red arrow in Figs.~\ref{fig:ROI1} and \ref{fig:ROI2}) can be seen in the FBP30NI and VEO30NI images, the blocky and pixelated effects in image appearance make them unacceptable for diagnostic use. The denoised images by CNN-VGG31 provide the best delineation of lesions relative to the ground truth of VEO10NI, while improving overall image appearance, which may greatly improve the diagnostic confidence.

\begin{table}[tb]
	\centering
	\caption{PSNR and SSIM of the denoised images.}
	\label{tab:psnr}
	\begin{tabular}{c|c|c|c|c|c|c}
		\hline\hline
		& FBP30NI &  CNN-MSE & CNN-VGG11 & CNN-VGG31 & CNN-VGG34 & BM3D \\
		\hline
		PSNR & 27.1544  & 31.1135 & \bf{31.1239} & 30.6462 & 30.2154 & 28.7405\\
		\hline
		SSIM & 0.8018 & \bf{0.9351} & 0.9348 & 0.9260 & 0.9159 & 0.9026\\
		\hline\hline
	\end{tabular}
\end{table}

The traditional metrics of PSNR and SSIM were also used for evaluation as shown in Table~\ref{tab:psnr}. PSNR is equivalent to the per-pixel loss. As measured by PSNR, a model trained to minimize per-pixel loss should always outperform a model trained to minimize feature reconstruction loss. 
Thus, it is not surprising that CNN-MSE achieves higher PSNR and SSIM than CNN-VGG31 and CNN-VGG34. However, these quantitative values are close, and the results of CNN-VGG31 and CNN-VGG34 are visually much more appealing. Overall, these two networks are better than CNN-MSE and CNN-VGG11.


In our experiments, we tested three feature maps of the VGG network. Generally speaking, lower-level layers of VGG extract primitive features, while higher-level layers give more sophisticated higher level features. This explains why CNN-VGG11 has a similar visual effect as CNN-MSE while CNN-VGG31 and CNN-VGG34 preserve more details. 

As for the computational cost, it took about 16 hours to train the CNN-MSE network and 10 minutes to fine-tune the CNN-VGG networks on a GTX980 GPU. After the networks were trained, restoring a single image took less than 5 seconds. Thus, compared with the typical time of CT image reconstruction, computational cost would never be a problem for image denoising using deep neural networks in clinical applications.

\section{Conclusions}
\label{sec:conclusions}

In this work, we have proposed a convolutional neural network for CT image denoising with a perceptual loss measure, which is defined as the MSE between the feature maps of the CNN output and the ground truth respectively. The experimental results show that the proposed network increases the images' PSNR and SSIM and that the perceptual regularization helps prevent image from over-smoothing and losing structure details. In our future work, we will refine, validate, and optimize our perceptive CNN with a larger dataset. More importantly, we will perform a reader study to compare the radiological reading reports with our deep learning results.


\bibliographystyle{splncs03}
\bibliography{denoisingCT}

\end{document}